\documentclass[
]{elsarticle}

\usepackage{lineno,hyperref}
\modulolinenumbers[5]

\journal{Pattern Recognition (accepted)}

\usepackage[rgb]{xcolor}
\usepackage{multirow}

\usepackage{amsmath,graphicx,amsfonts,amssymb}
\usepackage{dsfont}
\usepackage{pgf,tikz,pgfplots,pgfplotstable}
\usepackage{algorithm}
\usepackage{algpseudocode}

\pgfplotsset{compat=1.15}

\usepackage{eucal}

\usepackage{import}
\usetikzlibrary{3d} 

\pgfplotsset{
	wafer_map_style/.style={
		width=0.17\columnwidth,
		height=0.17\columnwidth,
		scale only axis,
		axis line style={draw=none},
		tick style = {draw = none},
		xticklabels = {},
		yticklabels = {},
		xmin=-1.05,
		xmax=1.05,
		ymin = -1.05,
		ymax = 1.05,
		title style = {font = \tiny, at={(0.5, 0.78)}, align=center, overlay},
	},
	big_wafer_map_style/.style={
		width=0.9\columnwidth,
		height=0.9\columnwidth,
		scale only axis,
		axis line style={draw=none},
		tick style = {draw = none},
		xticklabels = {},
		yticklabels = {},
		xmin=-1.05,
		xmax=1.05,
		ymin = -1.05,
		ymax = 1.05,
		title style = {font = \tiny, at={(0.5, 0.78)}, align=center, overlay},
	},
	defect_style/.style={
		color=blue,
		only marks,
		mark size = 0.2pt,
		mark = *,
		line width=0.5pt,
	}, 
	border_style/.style={
		line width=1pt,	
	}, 
}

\begin{document}
\begin{frontmatter}
\title{Deep Open-Set Recognition for Silicon Wafer Production Monitoring}

\author[mymainaddress]{Luca Frittoli\corref{mycorrespondingauthor}}
\cortext[mycorrespondingauthor]{DOI: 10.1016/j.patcog.2021.108488}
\ead{luca.frittoli@polimi.it}

\author[mysecondaryaddress]{Diego Carrera}
\author[mysecondaryaddress]{Beatrice Rossi}
\author[mysecondaryaddress]{Pasqualina Fragneto}
\author[mymainaddress]{Giacomo Boracchi}

\address[mymainaddress]{DEIB, Politecnico di Milano, via Ponzio 34/5, Milan (Italy)}
\address[mysecondaryaddress]{STMicroelectronics, via Camillo Olivetti 2, Agrate Brianza (Italy)}

\begin{abstract}
The chips contained in any electronic device are manufactured over circular silicon wafers, which are monitored by inspection machines at different production stages. Inspection machines detect and locate any defect within the wafer and return a Wafer Defect Map (WDM), i.e., a list of the coordinates where defects lie, which can be considered a huge, sparse, and binary image. In normal conditions, wafers exhibit a small number of randomly distributed defects, while defects grouped in specific patterns might indicate known or novel categories of failures in the production line. Needless to say, a primary concern of semiconductor industries is to identify these patterns and intervene as soon as possible to restore normal production conditions. 

Here we address WDM monitoring as an open-set recognition problem to accurately classify WDM in known categories and promptly detect novel patterns. In particular, we propose a comprehensive pipeline for wafer monitoring based on a Submanifold Sparse Convolutional Network, a deep architecture designed to process sparse data at an arbitrary resolution, which is trained on the known classes. To detect novelties, we define an outlier detector based on a Gaussian Mixture Model fitted on the latent representation of the classifier. Our experiments on a real dataset of WDMs show that directly processing full-resolution WDMs by Submanifold Sparse Convolutions yields superior classification performance on known classes than traditional Convolutional Neural Networks, which require a preliminary binning to reduce the size of the binary images representing WDMs. Moreover, our solution outperforms state-of-the-art open-set recognition solutions in detecting novelties.
\end{abstract}

\begin{keyword}
pattern classification \sep open-set recognition \sep sparse convolutions \sep quality inspection \sep wafer monitoring
\end{keyword}

\end{frontmatter}

\section{Introduction}\label{sec:intro}
\begin{figure*}[t!]
	\centering	
	\includegraphics[width=\textwidth]{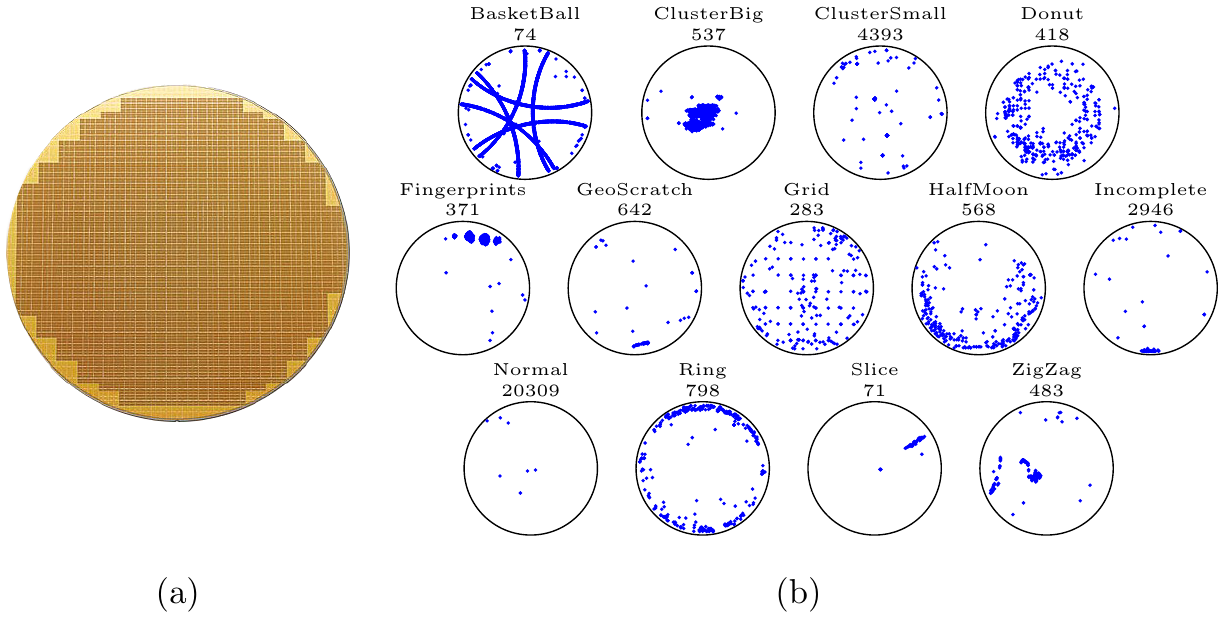}
	
	\caption{(a) An example of a wafer containing few hundreds of chips (the small cells). (b) An example of WDM for each known class in the ST dataset. We also report the number of instances of each class to emphasize the severe class imbalance in the ST dataset.}
	\label{fig:classes}
\end{figure*}

Silicon wafers are the first production stage of many electronic components, including processors, memories and sensors, that are present in any electronic device from smartphones to cars. Producing wafers requires costly, long, and high-tech industrial processes. In the last few decades, the demand and production volumes have steadily been growing, making manual quality inspection inadequate. Each wafer (Figure~\ref{fig:classes}(a)) contains hundreds of chips and has to be analyzed by several \emph{inspection machines} at different stages of the production process to locate any defect. Each wafer inspection outputs a Wafer Defect Map (WDM), namely a list of coordinates of the wafer (i.e., a 2-dimensional point cloud) where defects were found. In normal production conditions, defects are rare and randomly distributed in WDMs. In contrast, WDMs containing \emph{patterns} like those shown in Figure~\ref{fig:classes}(b) might be symptoms of problems or failures in the production line, which must be promptly classified to solve production failures as soon as possible, preventing the waste of time and resources. These patterns are known to be related to specific problems in a particular manufacturing step. Moreover, WDMs might exhibit novel patterns related to unknown production issues, and it is a primary concern of production engineers to detect also these. Detecting novel patterns in WDMs is perhaps a more challenging problem than classifying known patterns, which is of paramount importance for semiconductor companies. In this work, we address pattern classification on WDMs as an \emph{open-set recognition} problem \cite{scheirer2014probability}, to accurately classify WDMs corresponding to known categories, and at the same time detect novel patterns. Open-set recognition is a very active research area with applications in image classification \cite{geng2020guided, huang2020behavior} and face recognition \cite{dong2020open}, but has not been addressed in WDM monitoring yet. 

Pattern recognition on Wafer Defect Maps has been widely investigated \cite{huang2015automated}, but exclusively in a traditional, closed-set classification framework, where test instances have to be associated with a set of known classes. Even in a closed-set scenario, this is a challenging classification problem, due to the size of images representing WDMs, which is limited only by the resolution of the inspection machines. These images can be huge: in our case, the coordinates of defects span a grid of dimensions $20,000\times 20,000$, corresponding to a precision of 10 $\mu m$. Some inspection machines might achieve even higher precision. Hence, processing an entire WDM as a binary image would be impossible for a standard classifier (e.g., a Convolutional Neural Network) due to the large demand for memory and computational resources. For this reason, all the existing solutions transform WDMs into smaller images called \emph{Wafer Bin Maps}. This corresponds to a lossy conversion of the original WDM, which we believe might overlook important information, yielding worse classification performance. Handling WDMs at their original resolution is challenging and requires ad-hoc models, as is often the case in industrial applications where patterns have to be identified in point clouds \cite{bergamasco2020cylinders} or low-resolution/noisy images \cite{chen2017hybrid,abiantun2019ssr2}, to name a few examples.


In this paper we consolidate and extend our preliminary work \cite{di2019wafer} on WDM classification. Here we propose an effective and efficient pipeline for wafer production monitoring, addressing open-set recognition on full-resolution WDMs. Our solution is based on a Submanifold Sparse Convolutional Network (SSCN) \cite{graham20183d}, a model specifically designed to efficiently process sparse images, regardless of their resolution, as lists of their non-zero locations. 
Our main contributions are:
\begin{itemize}
    \item We present the first custom SSCN architecture that enables classifying patterns from full-resolution WDMs into a set of known classes.
    \item We are the first to address WDM monitoring as an open-set recognition problem. In particular, we propose to detect novelties from the latent representation of our SSCN, trained on known classes, by means of a Gaussian Mixture Model (GMM).
    \item We design a specific data augmentation procedure for WDMs to reduce the impact of class imbalance during training. Moreover, we show that augmentation can safely be used at test time also in an open-set scenario since it cannot transform novelties into instances of known classes.
\end{itemize}

In \cite{di2019wafer} we introduced Submanifold Sparse Convolutional Networks for WDM classification in a traditional, closed-set scenario, where all instances in the test set are assumed to belong to a set of known classes (Figure~\ref{fig:classes}(b)). Here, we modify our SSCN to address an open-set recognition problem, which is of paramount importance when monitoring wafer production. To assess the novelty-detection performance of our open-set classifier, we design a leave-one-out testing procedure that complies with industrial scenarios like ours, where annotated novelties are not readily available. We also investigate more in depth the data augmentation procedure proposed in \cite{di2019wafer}, and we experimentally show its superiority compared to traditional augmentation techniques in closed-set classification. Moreover, we demonstrate that our data augmentation can be safely used in an open-set scenario since it cannot turn a WDM containing a novel defect pattern into an instance of any known class.

Our experiments on the ST dataset, acquired and annotated at the STMicroelectronics plant in Agrate Brianza, Italy, demonstrate the advantages of handling WDMs directly as lists of coordinates, rather than converting them into low-resolution images. In fact, our SSCN achieves better classification performance over known classes than traditional CNNs trained on Wafer Bin Maps, despite having substantially fewer parameters. Moreover, our open-set recognition solution can detect novelties significantly better than alternatives from the literature, which we applied to our SSCN to enable a fair comparison.


The paper is organized as follows: in Section \ref{sec:related} we present the most recent works addressing wafer monitoring and open-set recognition. In Section \ref{sec:pf} we formally define the open-set recognition problem we address on WDMs. In Section \ref{sec:ps} we describe our wafer monitoring solution in detail. In Section \ref{sec:experiments} we present the experiments we perform to support our claims. Section \ref{sec:conclusion} concludes the paper with some final remarks and hints on ongoing and future works.

\section{Related work}\label{sec:related}
Several automatic methods to monitor the quality of the chips produced by semiconductor companies have been developed in the last few years. Typically, these methods consist of closed-set classifiers trained to recognize a set of known classes of either local defect in images of small wafers portions, acquired by an electronic microscope, or global defect patterns in Wafer Defect Maps: the interested reader can find a comprehensive survey in \cite{huang2015automated}. Here we survey the literature regarding defect analysis in wafers (Section \ref{subsec:waferPR}) and open-set recognition (Section \ref{subsec:openset}), which is the problem we address for wafer monitoring.

\subsection{Wafer monitoring}
\label{subsec:waferPR}
During production, wafers get inspected to identify localized defects \cite{sakata2005successive, li2012wavelet, cheon2019convolutional} and defect patterns in WDMs \cite{di2019wafer, yu2016wafer, fan2016wafer, chang2012development, nakata2017comprehensive, nakazawa2018wafer, yu2019wafer, saqlain2020deep}, which is the problem we address in our work. Wafer Defect Maps are lists containing the coordinates at which inspection machines find defects, and these correspond to huge binary images (in our case $20,000\times 20,000$). This makes WDMs a challenging kind of data to handle, impossible to feed to a standard CNN. For this reason, the vast majority of existing solutions preprocess WDMs to reduce their size, usually building Wafer Bin Maps, which are smaller images (say $200\times200$) where pixels are associated with small portions of the wafer, and the value of each pixel indicates whether the corresponding wafer portion contains defects or not. 

The first supervised methods \cite{yu2016wafer,fan2016wafer,chang2012development} employ hand-crafted features, including geometric regional features such as area, perimeter, or eccentricity of defect clusters \cite{yu2016wafer}, and density-based features such as the location of defect-dense areas \cite{fan2016wafer}. Other informative features can be obtained by analyzing Wafer Bin Maps in a different domain using the Radon or Hough transforms to highlight specific patterns \cite{chang2012development}. A few of these features are typically stacked in vectors and fed to a classifier such as a support vector machine (SVM) or a decision tree. However, hand-crafted features might not identify meaningful patterns in every condition, e.g., when patterns are rotated, shifted, or cover only part of the WDM. Moreover, hand-crafted features are usually defined to highlight defect patterns belonging to known classes. Thus, they might be meaningless for the detection of novel patterns.

Since Convolutional Neural Networks have achieved impressive results in image classification, the most recent methods \cite{nakata2017comprehensive,nakazawa2018wafer,yu2019wafer,saqlain2020deep} employ Deep Learning models to classify Wafer Bin Maps. In particular, \cite{nakata2017comprehensive} addresses the simplified problem of distinguishing radial map patterns from non-radial ones. The solution presented in~\cite{nakazawa2018wafer} proposes a specific preprocessing where the intensity value of each pixel represents the number of defects in the corresponding wafer portion. Our previous work \cite{di2019wafer}, and others \cite{yu2019wafer,saqlain2020deep} show the superiority of deep CNNs over traditional machine-learning methods based on hand-crafted features over the public dataset WM-811K \cite{wu2015wafer}, where they achieve excellent classification performance. However, in these works, wafer monitoring is tackled as a closed-set classification problem, so these methods cannot detect novel defect patterns. Moreover, the WM-811K dataset contains small images representing Wafer Bin Maps, so we cannot use it to test the proposed solution, which takes the original WDMs as input.


\subsection{Open-set recognition}
\label{subsec:openset}
Open-set recognition, introduced by Scheirer et al. \cite{scheirer2014probability}, is the problem of recognizing a certain number of known classes -- for which a large annotated dataset is available -- and detecting as \emph{novelties} those samples that do not belong to any known class. Compared to the traditional closed-set classification, where all the classes occurring at test time are assumed to be known, the open-set recognition problem refers to a more realistic scenario, where only part of the classes have been already identified and included in the training set.


The first open-set recognition methods are based on traditional machine-learning algorithms. Scheirer et al. \cite{scheirer2014probability} use modified Support Vector Machines with decision boundaries designed to reject unknown samples. Other methods detect novelties using the distance of test samples from the centroids of the known classes \cite{bendale2015towards}, or the reconstruction error of sparse representations \cite{zhang2016sparse}.

More recently, deep open-set recognition methods started to gain more and more attention due to the outstanding results achieved by deep learning in most classification and pattern-recognition tasks. Bendale et al. \cite{bendale2016towards} propose the OpenMax function to replace SoftMax as the last layer of a CNN at test time. In particular, a standard CNN is trained on the known classes and, during testing, OpenMax evaluates the distance between the CNN score vectors and the \emph{mean activation vectors} (MAVs), computed using the score vectors of training samples. Each MAV represents a known class, and a test sample is detected as a novelty when the likelihood of its distance from all the MAVs with respect to a Weibull distribution model fitted using the training set is below a certain threshold. Cevikalp et al. \cite{cevikalp2021deep} propose deep classifiers using polyhedral conic boundaries to separate instances from different known classes, instead of the traditional linear boundaries. This makes the acceptance regions of the known classes more compact, thus easing novelty detection.

Another approach consists in applying an outlier-detection method to the latent representation of a deep classifier trained on known classes \cite{zhu2018multi, zhang2020large}. Instances from known classes are expected to be mapped in the same region of the latent space, following a multimodal distribution. Therefore, it is possible to detect novelties from their latent representations as outliers with respect to this distribution. To this purpose, Zhu et al. \cite{zhu2018multi} employ a variant of Isolation Forest \cite{liu2008isolation}, while Zhang et al. \cite{zhang2020large} define confidence intervals over each component of the latent space. In the context of wafer monitoring, Cheon et al. \cite{cheon2019convolutional} follow the same approach to address open-set recognition on wafer surface defect images acquired by a scanning electron microscope (SEM). In particular, they apply a $k$-nearest neighbors outlier detector to the features extracted by a CNN trained on the known classes. However, this solution is designed to process traditional images of localized defects rather than WDMs, and thus it cannot be applied directly to the problem we address in this paper. 

Socher et al. \cite{socher2013zero} obtain a different latent representation by embedding the images into a semantic word space associated with the class labels. Then, they fit an isometric Gaussian model to represent each known class in the semantic space and use the likelihood as novelty score. More sophisticated deep methods for open-set recognition have been proposed, combining the reconstruction error of an autoencoder with classification features to detect novelties \cite{yoshihashi2019classification, oza2019c2ae, sun2020conditional}. However, these approaches based on autoencoders are designed for relatively small images and cannot be applied directly to WDMs. 

\section{Problem Formulation}\label{sec:pf}

A WDM $w$ is a list containing the 2-dimensional coordinates indicating where defects lie within a wafer. These coordinates belong to a huge grid defined by the resolution of the inspection machines. For pattern recognition purposes, $w$ can be seen as a binary image $w\in\{0,1\}^{K\times K}$, where each pixel $(i,j)$ corresponds to an inspected wafer location and $w(i,j) = 1$ if the coordinates $(i,j)$ belong to the list and 0 otherwise. In normal production conditions, defects are rare and randomly distributed within the wafer, whilst failures might cause defects arranged in patterns like those shown in Figure \ref{fig:classes}(b). These patterns might either belong to a known class $\ell\in\mathcal L$ indicating a known production problem, or indicate a novel issue that had never been observed before. Although these previously unseen instances are usually referred to as \emph{unknown} \cite{sun2020conditional} or \emph{out-of-distribution} \cite{hendrycks2016baseline} in the open-set recognition literature, we indicate them as \emph{novelties} to highlight that these correspond to new defect patterns. 

Our goal is to train an open-set classifier $\mathcal K$ that associates to each WDM $w$ either a known class label or the \emph{Novel} label, which identifies WDMs containing novel types of patterns, i.e., the output of $\mathcal K$ is:
\begin{equation}\label{eq:openclassifier}
    \mathcal K(w) = \begin{cases}
        \textit{Novel}\\
        \widehat{\ell}(w)\in\mathcal L.
    \end{cases}
\end{equation}
To this aim, we assume that a training set formed by $n$ annotated WDMs $\mathcal W=\{(w_1,\ell_1),\dots,(w_n,\ell_n)\}$ -- where each $\ell_i\in\mathcal L$ is a known label -- is provided. 

A major challenge of handling WDMs is that state-of-the-art deep learning methods such as Convolutional Neural Networks cannot be directly applied due to the huge size of the images obtained from full-resolution WDMs: in our case, $K=20,000$, so a WDM would require almost $3~$GB to be loaded in memory in single precision as a grey-scale image. A second challenge is the extreme class imbalance: indeed, the vast majority of WDMs belong to the \emph{Normal} class, while some patterns, such as \emph{BasketBall}, occur very rarely and are under-represented in the ST dataset, as shown in Figure~\ref{fig:classes}(b). 



\section{Proposed Solution}\label{sec:ps}

We address WDM monitoring as an open-set recognition problem and propose a network architecture to classify very efficiently full-resolution WDMs belonging to the classes represented in the training set (Section \ref{subsec:closed-set}). Then, we extend our network -- which is trained to address a traditional multi-class classification problem -- to open-set recognition, i.e., to detect WDMs containing novel patterns (Section \ref{subsec:open-set}). Here, we also describe the class-specific data augmentation procedure we employ in our open-set recognition method, both at training and test time (Section \ref{subsec:augmentation}) and summarize the proposed pipeline to classify WDMs (Section \ref{subsec:testphase}).

\subsection{Classification of known classes}\label{subsec:closed-set}
\begin{figure}[t!]
	\centering
	\includegraphics[width=\columnwidth]{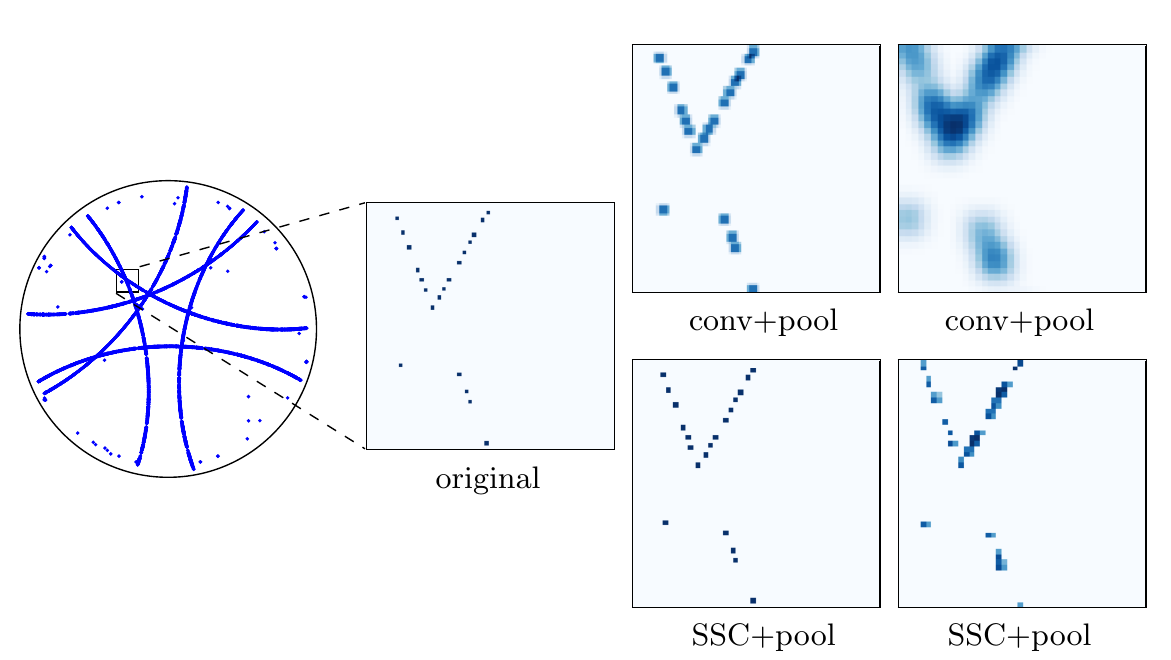}
		
	\caption{Visual representation of the difference between processing a WDM patch ($200\times200$) by regular convolutions (conv) and Submanifold Sparse Convolutions (SSC). All these convolutional layers have uniform filters of size $7\times7$ and are followed by max-pooling with a kernel of size $2\times2$, yielding a downsampling factor 2 on each dimension. For illustration purposes, we represent all the images at the same size. While regular convolutions reduce the original sparsity, SSCs maintain it since they do not increase the number of active locations.} 
	\label{fig:sparsity}
	\vspace{-0.2cm}
\end{figure}

The open-set recognition problem includes a traditional multi-class classification problem with a fixed set of known classes, i.e., those represented in the training set. Traditional Convolutional Neural Networks, which represent state of the art in image classification, cannot be directly applied to WDMs because they take as input relatively small images (e.g., VGG16~\cite{simonyan2014very} and ResNet50~\cite{he2016deep} take as input $224 \times 224$ RGB images), while images representing full-resolution WDMs are huge and result impossible to use to train and test CNNs. 

To handle WDMs efficiently, we build a deep network based on Submanifold Sparse Convolution (SSC)~\cite{graham20183d}, a modified convolutional operator designed to process sparse images at arbitrary resolution. The output of an SSC is the same as that of a regular convolution, but only on the active sites of its receptive field, namely, the non-zero locations, i.e.:
\begin{equation}
    \text{SSC}(w) = \text{conv}(w) \odot \mathds{1}(\text{supp}(w)),
    \label{eq:sparseconv}
\end{equation}
where $\odot$ indicates the Hadamard product, $\mathds{1}$ the indicator function and $\text{supp}(w)$ the support of $w$, i.e., the set of its non-zero locations. The main advantage of SSC compared to traditional convolution is that it enables a very efficient processing of sparse images, regardless of their resolution. Thus, a WDM can be processed directly as the list of the coordinates where defects lie within the wafer, which correspond to the active sites. Moreover, \eqref{eq:sparseconv} implies that SSC maintains the input sparsity -- thus the shape of the defective patterns in WDMs -- throughout the layers and does not increase the number of active sites, while regular convolutions reduce the sparsity of WDMs, as illustrated in Figure \ref{fig:sparsity}.

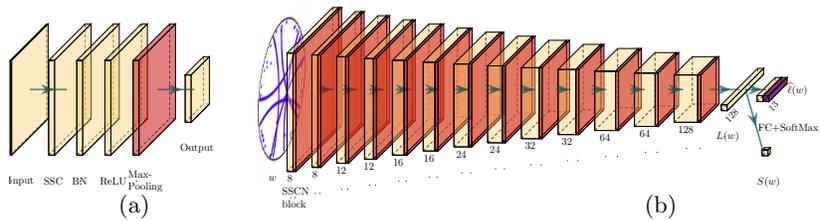
\begin{figure}[t!]
	\centering
	\includegraphics[width=\textwidth]{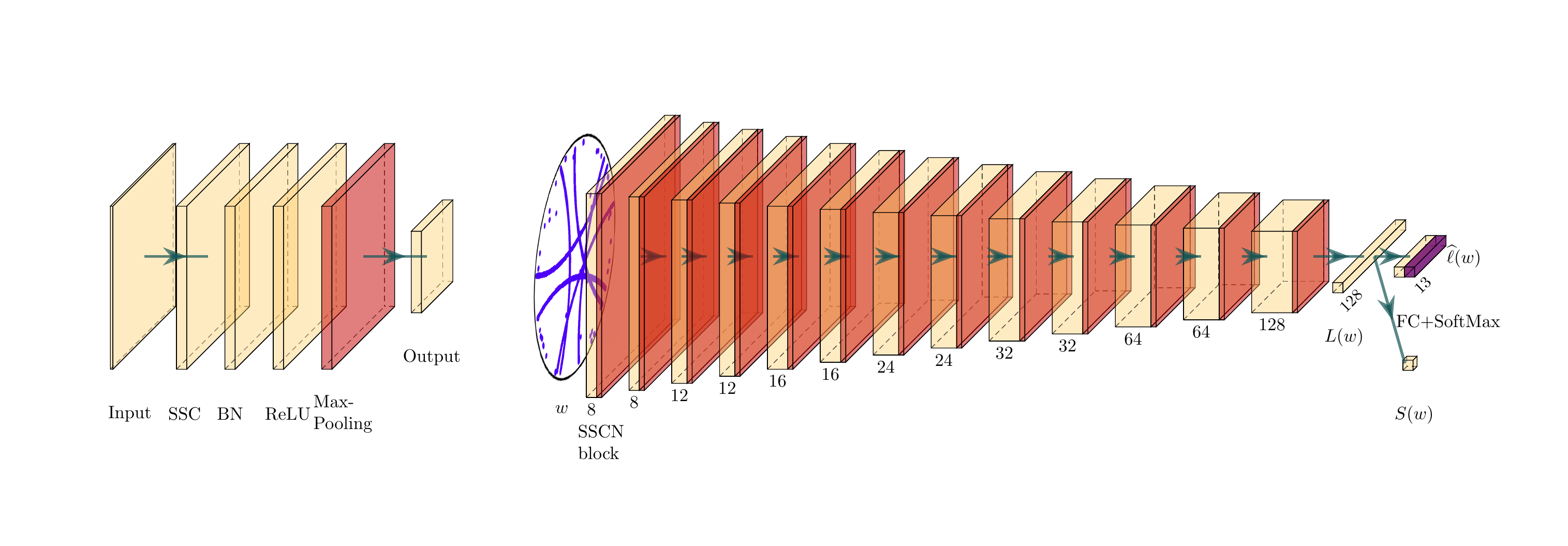}
	\begin{tikzpicture}[overlay]
	    \node at (-3.5,1.1) {\small (a)};
	    \node at (3.5,1.1) {\small (b)};
	\end{tikzpicture}
	\vspace{-1.2cm}
	\caption{(a): our SSCN block made of a Submanifold Sparse Convolutional layer (SSC), Batch Normalization (BN), ReLU activations and Max-Pooling layer (Pool) with stride 2. (b): our SSCN architecture, based on our SSCN block, for open-set recognition on WDMs. After 13 SSCN blocks, each WDM $w$ is transformed to a 128-dimensional latent representation $L(w)$, which is fed to a fully connected-layer (FC) with SoftMax activation to identify the known classes, and to the negative log-likelihood function of a GMM (fitted on the latent representations of WDMs from known classes), which is our novelty score $S(w)$.}
	\label{fig:arch}
\end{figure}

The architecture of the proposed SSCN is inspired by the convolutional part of the VGG16~\cite{simonyan2014very}, and is made of consecutive building blocks reducing the spatial resolution of the activation maps. Each block (Figure~\ref{fig:arch}(a)) includes an SSC layer, with Batch Normalization (BN) and ReLu activations, followed by a Max-Pooling layer of stride 2. Thus, a single block reduces the resolution of the feature map by a factor 2 on each dimension. Our architecture, illustrated by Figure~\ref{fig:arch}(b), is formed by 13 such blocks followed by a convolutional layer, and yields a 128-dimensional latent representation $L(w)$ of each WDM $w$, which we also employ to detect novelties (Section \ref{subsec:open-set}). Eventually, a fully-connected layer with SoftMax activations outputs a vector of $\# \mathcal L$ scores, whose maximum determines the predicted class of the input WDM.

We remark that both our solution and those based on traditional CNNs extract a compact representation of the WDMs \cite{nakazawa2018wafer, yu2019wafer, saqlain2020deep}. The main difference is that traditional CNNs for WDM classification require a preliminary binning of the WDMs to reduce their resolution, which we believe might lead to a loss of information. In contrast, our solution is entirely data-driven and does not discard any piece of information contained in WDMs, thanks to Submanifold Sparse Convolutions. Our SSCN yields a spatial downsampling factor $2^{13}$ on each dimension, way larger than what is customarily obtained by CNNs for image classification: for instance, the CNN proposed in \cite{saqlain2020deep} for Wafer Bin Maps only achieves a downsampling factor $2^5$. Although the network architecture made of convolutional and pooling layers recalls the VGG16~\cite{simonyan2014very}, our SSCN has a substantially fewer trainable parameters ($164,077$) due to the fact that the VGG16 has 3 large fully-connected layers on top of the convolutional part, which we do not include in our SSCN.

\subsection{Detection of novel patterns}\label{subsec:open-set}
\begin{figure}[t!]
    \centering	
    \includegraphics[width=\columnwidth]{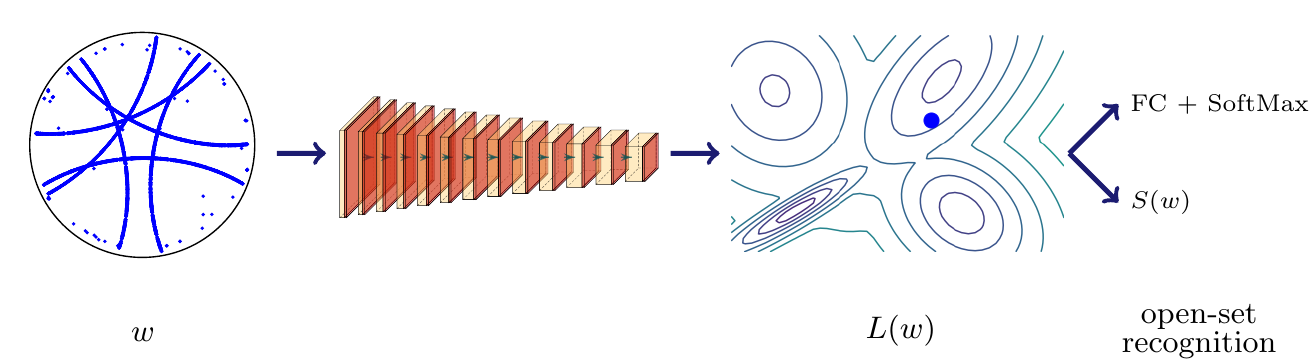}
    \caption{Visual representation of our approach to open-set recognition: we train a classifier on the known classes (SSCN), and detect as novelties those instances $w$ whose latent representation $L(w)$ through the SSCN are outliers in relation to the latent representation of the known classes. In particular, our novelty score $S(w)$ is the negative log-likelihood of a Gaussian Mixture Model $\widehat{\phi}$, represented in the image by its level curves.}
    \label{fig:approach}
\end{figure}
The second task in open-set recognition is detecting instances that do not belong to any known class, namely, novelties. This is particularly important in monitoring WDMs because a novel class of defect patterns must be detected as soon as possible and studied to determine which failure caused that unknown pattern. We follow the simple but effective idea illustrated in Figure \ref{fig:approach} and apply an outlier detector to the latent representation of our classifier -- i.e., the output of the penultimate layer of the network. This approach leverages the fact that a classifier maps instances belonging to the same class in the same area of the latent space so that the last, fully connected layer can separate the classes by linear boundaries \cite{zhu2018multi,zhang2020large}. Hence, the latent representation of the known classes can be described by a multimodal distribution $\phi$, where each mode is associated with a known class. We expect novelties to be mapped in low-density regions of the latent space with respect to $\phi$. For this reason, we fit a Gaussian Mixture Model (GMM) with $\#\mathcal L$ components, namely the number of known classes, and obtain a density $\widehat{\phi}$ to describe the distribution of the latent representations $L(w)$ extracted by the SSCN trained on WDMs from known classes, i.e.:
\begin{equation}
    \widehat{\phi} = \sum_{i=1}^{\#\mathcal L}\widehat{a}_i\cdot \mathcal N(\widehat{\mu}_i, \widehat{\Sigma}_i),
\end{equation}
where each Gaussian component $\mathcal N(\widehat{\mu}_i, \widehat{\Sigma}_i)$ for $i=1,\ldots,\#\mathcal L$ represents a known class. Then, at test time, we employ as novelty score the negative log-likelihood of $\widehat{\phi}$, i.e., for each WDM $w$,
\begin{equation}\label{eq:noveltyscore}
S(w) = -\log\big(\widehat{\phi}(L(w))\big).
\end{equation}
Eventually, a WDM $w$ is detected as \emph{Novel} when $S(w)>\eta$, where $\eta$ is a threshold we set such that $\mathbb{P}(S(w)\geq \eta\;|\;\ell\in\mathcal L)=\alpha$, where $\ell$ is the label associated to $w$ and $\alpha$ is the target false positive probability. 

To prevent $\widehat{\phi}$ from overfitting the latent representation of the training set, we use 90\% of the data to train the SSCN, and compute the latent representation of another 5\% of the training set. Then, we use these latent representations to fit $\widehat{\phi}$ by estimating its parameters using the well-known \emph{Expectation Maximization} algorithm \cite{dempster1977maximum}. Finally, we compute the novelty score $S(w)$ \eqref{eq:noveltyscore} of each instance $w$ of the remaining 5\% of the training data, and set the threshold $\eta$ as the empirical $(1-\alpha)$-quantile of these novelty scores. Since we employ the same latent representation for recognizing known and novel patterns, novelty detection does not require additional computations other than those needed to compute $S(w)$, compared to closed-set classification.



\subsection{Data augmentation}\label{subsec:augmentation}
The ST dataset is relatively small compared to the traditional datasets used for image classification and is characterized by an extreme class imbalance. Both these facts increase the risk of overfitting when training a deep classifier, and, in particular, the classification performance is likely to be poor on under-represented classes. To address this problem, we implement an ad-hoc data augmentation procedure based on a set of \emph{label-preserving} transformations to be applied to the WDMs.
Let $\mathcal T^{\ell}$ denote the set of label-preserving transformations for WDMs belonging to the class $\ell$: 
\begin{equation}\label{eq:def_transformations}
\mathcal T^{\ell} = \left\{T_{\boldsymbol\theta}^\ell\colon \{0,1\}^{K\times K}\to \{0,1\}^{K\times K},\;\boldsymbol{\theta}\in\Theta_\ell\right\},
\end{equation} 
where $\boldsymbol\theta$ indicates the parameters that define each transformation $T^\ell_{\boldsymbol\theta}$, and $\Theta_\ell$ is the set of transformation parameters specific for each class $\ell\in\mathcal L$. Each $T^\ell_{\boldsymbol\theta}$ combines different transformations widely used for data augmentation on images, such as rotations around the center, horizontal flips, and small translations of the defective coordinates. Together with this customary set of transformations, we also apply two transformations that we specifically designed for WDMs:

\noindent\textbf{Noise injection} consists in adding a small number of defects to each WDM at randomly sampled coordinates. This operation does not change the label of a WDM because a few randomly distributed defects are present in every wafer due to impurities naturally present in silicon. In particular, WDMs from the \emph{Normal} class can be seen as pure noise because the defects do not form patterns associated with specific problems in the production line but are few and randomly spread within the wafer, as can be expected when the process is executed normally. For this reason, we compute the empirical distribution $\widehat{\psi}$ of the number of defects that are present in \emph{Normal} WDMs in the ST dataset, and we draw from $\widehat{\psi}$ the number $D$ of defects to be injected in a WDM during augmentation. Our study and the experience of production engineers confirm that defects in \emph{Normal} WDMs do not show any specific pattern. Thus we inject $D$ defects at uniformly sampled polar coordinates in $[0,2\pi]\times[0,R]$, where $R$ is the wafer radius.

\noindent\textbf{Random mixing} consists in creating new training samples from under-represented classes, such as \emph{BasketBall} and \emph{Slice}, by superimposing randomly cropped parts of WDMs from the same class. This procedure is somewhat similar to \emph{mixup}~\cite{zhang2018mixup}, which builds new training samples as linear combinations of instances of the training set. However, mixup is designed for relatively small images and cannot be directly applied to WDMs. Another difference is that we do not change the label of a WDM after applying random mixing. To this purpose, we empirically verified that this process preserves the label: production engineers at STMicroelectronics could not tell the original WDMs from those generated by random mixing.

We execute this data augmentation procedure in each training epoch to produce new augmented batches, and also when fitting $\widehat{\phi}$, to obtain a sufficient number of samples from under-represented classes. This is done by generating several versions of each original WDM $w$ using class-specific transformations $T_{\boldsymbol\theta}^\ell(w)$, whose parameters $\boldsymbol{\theta}$ are randomly sampled from $\Theta_\ell$. 

\noindent\textbf{Test time augmentation.} To stabilize the output and improve the classification performance we employ data augmentation also at test time, averaging the classification scores obtained on different augmented versions of each WDM from the test set, as in \cite{simonyan2014very}. Indeed, even though the features extracted by the network should, in principle, be invariant to the label-preserving transformations in $\mathcal T^\ell$, perfect invariance cannot be achieved in practice, thus combining the predictions of several augmented versions of the same WDM typically improves the classification performance \cite{simonyan2014very}. Since the labels of the test set cannot be assumed to be known, we define the set $\mathcal T$ of transformations that preserve all the labels $\ell\in\mathcal L$:
\begin{equation}\label{eq:def_common_transformations}
\mathcal T = \bigg\{T_{\boldsymbol\theta}\,:\,{\boldsymbol\theta}\in\Theta=\bigcap_{\ell\in\mathcal L}\Theta_\ell\bigg\}.
\end{equation}
We exclude from $\mathcal T$ our random mixing transformation, which constructs new training samples from under-represented classes such as \emph{BasketBall} and \emph{Slice}.

To apply data augmentation when testing an open-set recognition method, it is important to verify that the transformations in $\mathcal T$ preserve not only the known class labels $\ell\in\mathcal L$, but also the \emph{Novel} class label considered at test time. This is certainly guaranteed when $\mathcal T$ is a group since any $T_{\boldsymbol\theta} \in\mathcal T$ that transforms a novelty into a sample of a known class would have an inverse $T_{\boldsymbol\theta}^{-1}\in\mathcal T$ that does not preserve the known class labels because it transforms an instance of a known class into a novelty. Here is a contradiction, therefore every $T_{\boldsymbol\theta} \in\mathcal T$ preserves the \emph{Novel} class label. 

In our case, $\mathcal T$ is not a group because noise injection has no inverse transformation in $\mathcal T$. However, noise injection naturally preserves every label, including \emph{Novel}, since it simulates the noise affecting all the manufactured wafers. We can obtain the other transformations $T_{\boldsymbol\theta} \in\mathcal T$ by composing, in any order, a rotation around the center of an angle $\beta\in\{0^{\circ}, 90^{\circ}, 180^{\circ}, 270^{\circ}\}$, an optional horizontal flip and a translation in a random direction with maximum distance $\nu$. The inverse $T_{\boldsymbol\theta}^{-1}$ of an element of $\mathcal T$ can be obtained by composing, in the opposite order, the inverses of the rotation, horizontal flip, and translation that define $T_{\boldsymbol\theta}$. This implies that also $T_{\boldsymbol\theta}^{-1}\in\mathcal T$, hence $\mathcal T$ preserves the \emph{Novel} class label by the same argument used above for the case in which $\mathcal T$ is a group. Thus, we can safely employ the transformations in $\mathcal T$ for data augmentation at test time without influencing the novelty-detection performance.

\subsection{WDM monitoring pipeline}\label{subsec:testphase}
Here we summarize the proposed pipeline to classify a test WDM $w$: first, we generate a set of $N$ augmented maps
\begin{equation}\label{eq:augmentedTest}
\mathcal A_w = \{w_i=T_{\boldsymbol \theta_i}(w),\; i=1,\dots,N\},
\end{equation}
where each $\boldsymbol\theta_i$ is randomly sampled from $\Theta$ (Section \ref{subsec:augmentation}). Then, we feed all the WDMs in $\mathcal A_w$ to the network, and average both the novelty score and the classification scores over all the augmented versions of $w$ contained in $\mathcal A_w$ to stabilize the output, as in \cite{simonyan2014very}. Thus, output of our classifier $\mathcal K$ is: 
\begin{equation}\label{eq:output}
    \mathcal K(w) = \begin{cases}
        \textit{Novel}\;\; \text{if }\frac{1}{N}\sum_{i=1}^N S(w_i) > \eta\\
        \widehat{\ell}=\arg\max_{\ell\in\mathcal L} \frac{1}{N}\sum_{i=1}^N \text{SSCN}(w_i)\;\; \text{otherwise},
    \end{cases}
\end{equation}
where $\eta$ is the threshold defined in Section \ref{subsec:open-set} for the novelty score $S$ and $\text{SSCN}(w_i)$ indicates the classification scores of our SSCN (Section \ref{subsec:closed-set}) obtained from $w_i\in\mathcal A_w$. Thus, when a WDM is not detected as \emph{Novel}, it is classified by taking the label $\ell\in\mathcal L$ maximizing the traditional classification score, averaged over the $N$ augmented versions of $w$ contained in $\mathcal A_w$.

\section{Experiments}\label{sec:experiments}
Our experiments show that: \emph{i}) Submanifold Sparse Convolutional Networks handling Wafer Defect Maps at full resolution outperform traditional CNNs trained on low-resolution Wafer Bin Maps, \emph{ii}) our data augmentation procedure is crucial to achieve good classification performance, and \emph{iii}) our open-set recognition solution based on a GMM fitted on the latent representations extracted by our SSCN can detect novel patterns better than state-of-the-art open-set recognition methods.

\subsection{Experimental Set Up}
\noindent \textbf{Dataset.} We test our solution on the \emph{ST dataset}, which contains 31,893 WDMs acquired at the STMicroelectronics plant in Agrate Brianza, Italy. These WDMs are either annotated as \emph{Normal}, i.e., they do not contain any defect patterns, or belong to one of the 12 defect classes identified by STMicroelectronics engineers, which are illustrated in Figure~\ref{fig:classes}(b). 

Open-set recognition performance is typically assessed over models trained on datasets with numerous classes such as CIFAR-100 \cite{krizhevsky2009learning}, and ImageNet \cite{russakovsky2015imagenet} so that a certain number of classes can be taken out during training and considered \emph{Novel} at test time. However, in our industrial scenario, we have only 12 defect classes, and WDMs containing novel patterns have not been included in the dataset. For this reason, we follow a leave-one-out approach by training our model on all the known classes except a single defect class, which is considered \emph{Novel} at test time. By doing so, we assess the capability of our model to recognize each defect class as novel when trained on the other classes. Since we want to detect novel defect patterns, we always consider the \emph{Normal} class as known.

\noindent \textbf{Figures of Merit.} We assess the accuracy of the proposed SSCN on each known class by the \emph{confusion matrix}. We also provide an overall evaluation of the classification performance using two multi-class extensions of the \emph{Area Under the ROC Curve} (AUC). The first one is the \emph{1vsRest}-AUC~\cite{provost2000well}, which is a weighted average of the AUC values obtained in all the binary classification problems where each class is selected in turns as the positive class and all the remaining classes are merged in the negative class (1vsRest). We remark that, since the weight given to each positive class is its frequency in the test set~\cite{provost2000well}, the 1vsRest-AUC is influenced by class proportions. The second one is the \emph{1vs1}-AUC~\cite{hand2001simple}, which is the average of the AUC values of the binary classification problems between every pair of classes (1vs1). Hand and Till \cite{hand2001simple} show that, contrarily to the 1vsRest-AUC, the 1vs1-AUC is not influenced by the class proportions in the test set. Since the ST dataset is extremely imbalanced, it is important to observe both these figures of merit to analyze the classification performance on under-represented classes and assess our data augmentation procedure. 

We employ 10-fold cross-validation and average all these figures of merit over the 10 test folds to provide an overall assessment of the classification performance. We also rank the considered methods (rank~$=1$ for the best method, 2 for the second-best, and so on) according to the 1vsRest-AUC and the 1vs1-AUC, and compute their average rank over the 10 test folds to compare their classification performance, as suggested in \cite{demvsar2006statistical}.

Detecting novel patterns is, in fact, a binary classification problem where novelties represent the positive class (\emph{Novel}), and the known classes used for training are merged in the negative class (\emph{Known}). Therefore, we can directly compute the AUC of these binary classification problems, which does not depend on class proportions, and this is crucial because novelties are rare. In contrast to other metrics such as F-scores, the AUC does not depend on how detection thresholds on the novelty scores are set, which allows to evaluate the effectiveness of different methods without setting the thresholds. 
To provide an overall assessment of the novelty-detection performance, we rank the considered methods according to the AUC and compute their average rank over the different novel classes as suggested in \cite{demvsar2006statistical}.

\subsection{Classification of known classes}\label{subsec:classification}

In this experiment, we evaluate the classification performance of our solution over the thirteen classes identified so far by STMicroelectronics engineers. We train our SSCN using the Adam optimizer \cite{kingma2014adam} on an Nvidia Titan Xp GPU. Training the model for 100 epochs requires about 8 hours, while the average time to classify a WDM is $0.061 \pm 0.055$ seconds. This time includes the generation and processing of $N=250$ augmented WDMs as in \eqref{eq:augmentedTest}. We explain the large variability of the classification time by observing that the number of operations executed by SSC layers directly depends on the input WDM sparsity, which varies substantially from class to class.

\begin{figure}[t!]
	\centering
	\includegraphics[width=0.8\columnwidth]{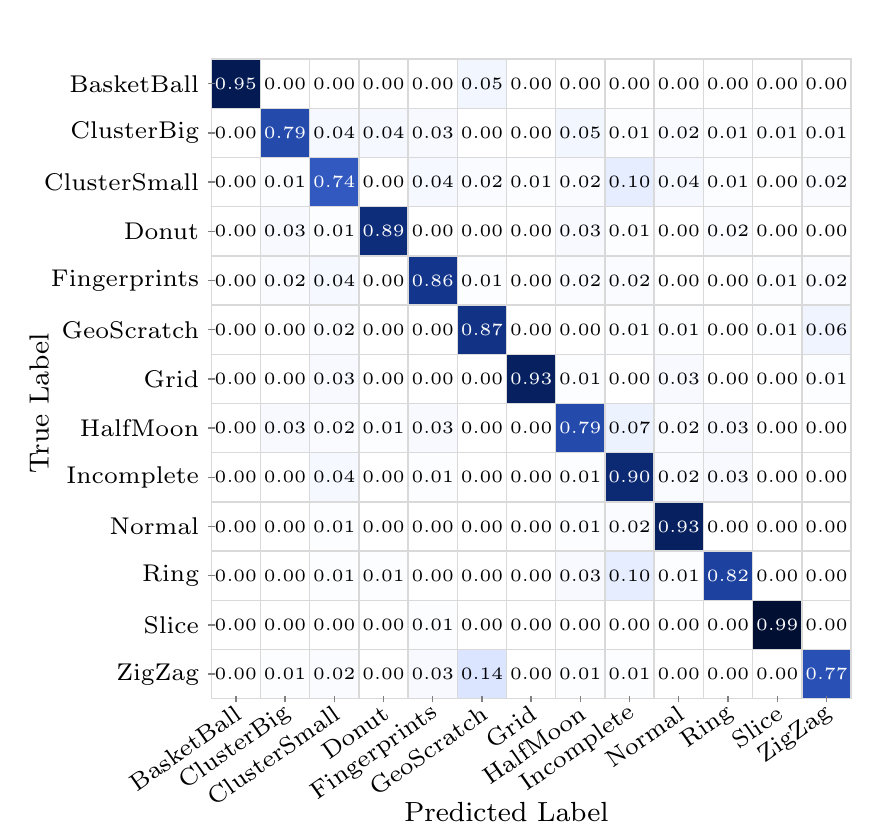}
	\vspace{-0.3cm}
	\caption{Confusion matrix of our SCNN obtained by 10-fold cross-validation on the ST dataset. Our network achieves very high accuracy on all the classes, and most misclassified samples belong to classes that are very similar, such as \emph{ClusterSmall} and \emph{Incomplete}.}
	\label{fig:conf:_mat}
	\vspace{-0.2cm}
\end{figure}

\noindent\textbf{Considered methods.} We compare the classification performance of the proposed SSCN with traditional CNNs for image classification. To this purpose, we reduce the resolution of the WDMs from $20,000\times20,000$ to $224\times 224$ by binning and converting each WDM to a grayscale image where the intensity of each pixel is the number of defects that fall into the corresponding bin. Then, we use the resulting Wafer Bin Maps to fine-tune the \emph{VGG16}~\cite{simonyan2014very} and \emph{ResNet50}~\cite{he2016deep} models, both pre-trained on the ImageNet dataset~\cite{russakovsky2015imagenet}. We fine-tune the VGG16 and ResNet50 using the Adadelta \cite{zeiler2012adadelta} and Adam \cite{kingma2014adam} optimizers, respectively, since these yield the best performance. 

We compare our SSCN to these models because fine-tuning CNNs for image classification is rather standard in wafer monitoring \cite{wu2015wafer}, because it reduces the risk of overfitting. We do not employ custom architectures for wafer classification \cite{yu2019wafer,saqlain2020deep} since the parameters of these models, pre-trained on the WM-811K dataset~\cite{wu2015wafer}, are not publicly available. Therefore, these deep models with several million parameters would risk overfitting when trained from scratch on the relatively small ST dataset. For a fair comparison with our SSCN, we train and test the VGG16 and ResNet50 using the same data augmentation described in Section \ref{subsec:augmentation} on the original WDMs, before reducing their resolution.

To assess the impact of data augmentation, we also evaluate the performance of our SSCN without data augmentation (both at training and test time), and with traditional augmentation, i.e., using only geometric transformations (translations, rotations, and flips). This experiment evaluates the importance of our custom transformations, namely noise injection and random mixing, in data augmentation.



\begin{table}[t!]
    \caption{Classification accuracy of the considered methods obtained by 10-fold cross-validation. We also report the average 1vsRest-AUC and 1vs1-AUC of the considered methods, their average rank on the 10 folds according to these metrics, and the p-values of the Wilcoxon test assessing the significance of the difference between the best-ranking method and each alternative \cite{demvsar2006statistical}. We also evaluate the accuracy, 1vsRest-AUC and 1vs1-AUC of our SSCN without augmentation and of all the considered methods with traditional augmentation, and report the differences between these and the corresponding performances achieved with our augmentation procedure. Here the Wilcoxon test assesses the significance of the performance difference compared to the same method with our augmentation procedure.}
	\label{tab:results_st}
	\centering
	\setlength{\tabcolsep}{9pt}
	\small
	\vspace{0.1cm}
	\resizebox{\textwidth}{!}{
	\begin{tabular}{l|ccc||c|ccc}
          & \multicolumn{3}{c||}{\scriptsize\textsc{our augmentation}} & \scriptsize\textsc{w/o aug.} & \multicolumn{3}{c}{\scriptsize\textsc{traditional augmentation}}\\
        Class accuracy & SSCN & VGG16 & ResNet50 & SSCN & SSCN & VGG16 & ResNet50\\
		\hline
		BasketBall	&        \textbf{0.9545} &	0.9091 &	0.9091 & $-0.7273$ &	$-0.0909$ & $-0.2727$ &	$-0.0909$\\
        ClusterBig	&        0.7880 &	0.7805 &	\textbf{0.7917}	& $-0.1013$ & $+0.0075$ &	$-0.0809$ &	$-0.0188$\\
        ClusterSmall	&    0.7439 &	\textbf{0.7844} &	0.7216	& $-0.0221$ & $+0.0143$ &	$-0.0436$ &	$+0.0492$\\
        Donut	&            \textbf{0.8933} &	0.8178 &	0.8133 & $-0.2311$ & $-0.0356$ & $-0.0221$ & $-0.0133$\\
        Fingerprints	&    \textbf{0.8571} &	0.7967 &	0.7857 & $-0.3764$ & $+0.0110$ & $-0.0773$ & $-0.0110$\\
        GeoScratch &	 0.8694 &	\textbf{0.8774} &	0.8678 & $-0.3264$ & $-0.0287$ & $-0.1284$ & $-0.0239$\\
        Grid	&            0.9261	& \textbf{0.9545} &	0.9375 & $-0.2841$ & $-0.0227$ & $-0.0352$ & $+0.0000$\\
        HalfMoon	&        \textbf{0.7912} &	0.7349 & 0.7470	& $-0.3855$ & $-0.0221$ & $-0.0064$ & $+0.0000$\\
        Incomplete	&        0.8965 &	\textbf{0.9097} &	0.8904 & $-0.1565$ & $-0.0136$ & $-0.0177$ & $-0.0037$\\
        Normal	&            0.9332 &	0.9132 & \textbf{0.9596} & $+0.0248$ & $-0.0045$ & $-0.0306$ & $-0.0093$\\
        Ring	&            0.8224 &	\textbf{0.8567} &	0.8552 & $-0.0806$ & $+0.0060$ & $-0.0026$ & $-0.0239$\\
        Slice	&            \textbf{0.9859} &	0.9437 &	0.9437 & $-0.4366$ & $-0.0141$ & $+0.0037$ & $+0.0000$\\
        ZigZag	&            \textbf{0.7681} &	0.7638 &	0.7064 & $-0.3277$ & $+0.0213$ & $-0.1472$ & $+0.0000$\\
        \hline
        1vsRest-AUC	&        0.9858 &	0.9818 &	\textbf{0.9868}	& $-0.0239$ & $-0.0007$ & $-0.0013$ & $-0.0019$\\
        Avg. rank	&        1.8000 &	3.0000 &	\textbf{1.2000} & & \\
        Wilcoxon p	&        0.0142 &	0.0025 &	\textbf{--}	& \textcolor{white}{$-$}0.0025 & \textcolor{white}{$-$}0.0844 &	\textcolor{white}{$-$}0.0844 &	\textcolor{white}{$-$}0.0372\\
        \hline
        1vs1-AUC	&        \textbf{0.9893} &	0.9846 &	0.9849	& $-0.0614$ & $-0.0018$ & $-0.0066$ & $-0.0029$\\
        Avg. rank	&        \textbf{1.0000} &	2.6000 &	2.4000 & & \\
        Wilcoxon p	&        \textbf{--} &	    0.0025 &	0.0035	& \textcolor{white}{$-$}0.0025 &	\textcolor{white}{$-$}0.0083 & \textcolor{white}{$-$}0.0025 & \textcolor{white}{$-$}0.0063\\
	\end{tabular}
	}
	\vspace{-0.3cm}

\end{table}

\noindent\textbf{Results and discussion.} Figure~\ref{fig:conf:_mat} shows the confusion matrix of our SSCN network obtained by 10-fold cross-validation on the ST dataset. Our model achieves excellent classification accuracy, and we observe that the majority of prediction errors occur when our model fails to distinguish two classes containing very similar patterns (e.g., \emph{ClusterSmall} and \emph{Incomplete}, see Figure~\ref{fig:classes}(b)). To make the comparison easier, we report in the first three columns of Table~\ref{tab:results_st} only the class accuracy (i.e., the diagonal of the confusion matrix) of our SSCN, VGG16, and ResNet50, all trained and tested using our augmentation procedure. Table~\ref{tab:results_st} also contains the average 1vsRest-AUC and 1vs1-AUC achieved by the considered methods on the test folds to compare their overall performance. As recommended in~\cite{demvsar2006statistical}, we also report the average rank of the considered method and the p-values of the one-sided Wilcoxon Signed-Rank test~\cite{wilcoxon1945individual} assessing whether the difference between the 1vsRest-AUC and the 1vs1-AUC of the best-ranking method and those of the other methods is significant. 

First, we observe that methods trained and tested with our data augmentation procedure achieve consistently high accuracy, and that our SSCN outperforms the other methods on six out of thirteen classes. Overall, our SSCN is the best in terms of 1vs1-AUC on all the 10 test folds (avg. rank$=1$), and the Wilcoxon Signed-Rank test confirms that the differences are statistically significant (p-value $\leq0.05$). Notably, our SSCN outperforms the VGG16, which has a similar architecture to our SSCN, in both 1vsRest-AUC and 1vs1-AUC. The ResNet50 model is the best in terms of 1vsRest-AUC thanks to its high accuracy on the \emph{Normal} class (approximately $96\%$), which is by far the most represented in the ST dataset (see Figure \ref{fig:classes}(b)), and the 1vsRest-AUC is sensitive to class proportions \cite{provost2000well}. 
%
Our SSCN provides a better trade-off between the accuracy on the \emph{Normal} class (over $93\%$) and accuracy on the other classes, as it outperforms the ResNet50 on nine out of twelve defect classes, often by a substantial margin. This explains the significant difference between the two in terms of 1vs1-AUC, which is not influenced by class proportions \cite{hand2001simple}.

To assess the impact of our data augmentation procedure on classification, we report in the fourth column of Table~\ref{tab:results_st} the performance differences between the proposed solution, which leverages data augmentation both during training and testing, and the same SSCN architecture trained and tested without augmentation. 
Remarkably, training the VGG16 and ResNet50 without data augmentation leads to overfitting due to the fact that the ST dataset is relatively small and the considered CNNs have several million trainable parameters. In contrast, our SSCN can be effectively trained without augmentation since it has a substantially lower number of parameters. However, without augmentation, our SSCN achieves substantially lower classification accuracy on all classes except the \emph{Normal} class, which is by far the most common in the ST dataset (see Figure \ref{fig:classes}(b)) and therefore can also be learned very accurately (approximately 96\%) without augmentation. The SSCN without augmentation also achieves lower 1vsRest-AUC and 1vs1-AUC compared to our SSCN trained using our data augmentation procedure, and the differences are statistically significant (p-value $\leq0.05$) according to the Wilcoxon test. 

Similarly, the last three columns of Table~\ref{tab:results_st} report the performance differences between the models trained and tested with our augmentation procedure (which includes noise injection and random mixing) and the same models trained and tested using only traditional augmentation, namely geometric transformations. 
We observe that traditional augmentation yields lower accuracy on most classes, and lower 1vsRest-AUC and 1vs1-AUC compared to the same model with our augmentation. According to the Wilcoxon test, the difference in 1vsRest-AUC is significant (p-value $\leq0.05$) for the ResNet50. Remarkably, the differences in 1vs1-AUC are significant for all the considered models, which confirms that our augmentation procedure improves the robustness to class imbalance, since the 1vs1-AUC is not influenced by class proportions \cite{hand2001simple}.

\subsection{Detection of novel patterns}
In this experiment, we assess the performance of our solution in detecting WDMs containing novel defect patterns. After training our SSCN on the known classes, we fit a GMM on the latent representation and use the log-likelihood as novelty score, as illustrated in Section \ref{subsec:open-set}. In particular, we train twelve different models, each time taking out one of the defect classes, which we then consider \emph{Novel} at test time. By doing so, we assess the open-set recognition performance of our solution in a realistic scenario where a new defective class unexpectedly occurs due to a production failure. Since some of the identified defect classes in the ST dataset (e.g., \emph{ClusterSmall} and \emph{Incomplete}) are very similar, we expect some novelties to be more difficult to detect than others.


\noindent\textbf{Considered methods.} We compare the novelty-detection performance of our solution against methods implementing the following open-set recognition techniques on the same SSCN:

\begin{itemize}
    \item \emph{SoftMax} is a widely employed open-set recognition baseline \cite{hendrycks2016baseline}. It is based on the intuition that novelties are typically classified with low confidence. Let $\textbf{v}$ be the score vector associated to a WDM $w$, i.e., the output of the last fully-connected layer of our SSCN (or any neural network). Then, the SoftMax is defined as:
    \begin{equation}
        \textbf{p}_j = \text{SoftMax}(\textbf{v})_j = \dfrac{\exp(\textbf{v}_j)}{\sum_{i=1}^{\#\mathcal L}\exp(\textbf{v}_i)},
    \end{equation}
    which guarantees that the scores are non-negative and sum to 1. The novelty score is simply $S(w)=-\max_j\textbf{p}_j$, which is high when the posterior probability of the selected class is low. 
    \item \emph{PreSoftMax} is based on the same idea as SoftMax, but uses as novelty score the opposite of the maximum classification score before applying SoftMax, i.e., $S(w)=-\max_j \textbf{v}_j$. As remarked in \cite{hendrycks2016baseline}, even if all the scores $\textbf{v}$ of a novel instance $w$ are low, SoftMax might still assign $w$ to one of the known classes with extremely high confidence since posteriors are forced to sum to 1, thus preventing SoftMax to detect $w$ as \emph{Novel}. The PreSoftMax scores are not subject to this constraint, so they might be more informative than the SoftMax scores in detecting novelties.
    \item \emph{OpenMax} is a function designed in \cite{bendale2016towards} to replace SoftMax in open-set classifiers. This method employs as novelty score the likelihood of a Weibull distribution fitted on the distances between the score vectors $\textbf{v}$ of a classifier and the Mean Activation Vectors (MAVs) of each known class, computed from the training set.
    \item We denote by \emph{SME} the open-set recognition method based on SoftMax Entropy proposed in \cite{geng2020guided}. SME relies of the observation that the posterior probability vectors of novel instances tend to have a larger Shannon entropy compared to instances of known classes, thus the novelty score is defined as the Shannon entropy of the posterior probabilities obtained by our SSCN, i.e., $S(w) = -\sum_j \textbf{p}_j \log \textbf{p}_j$. This method is part of a broader solution for open-set recognition and one-shot detection~\cite{geng2020guided}, but here we only consider the proposed open-set recognition solution.
    \item We denote by \emph{IFOR} the open-set recognition method proposed in \cite{zhu2018multi} that applies the Isolation Forest \cite{liu2008isolation} outlier detector to the latent representation of the WDMs, i.e., to $L(w)$. The method presented in \cite{zhu2018multi} considers a latent representation including the features extracted by a classifier and an embedding of the class labels to a semantic word space. However, in our industrial setting, there is no meaningful embedding of the labels to a semantic space. Hence we only consider $L(w)$ as input.
    \item We denote by \emph{CI} the open-set recognition method proposed in \cite{zhang2020large} that applies an outlier detector based on Confidence Intervals to the latent representation of the classifier, treating each component of $L(w)$ as an independent variable. This method computes a confidence interval $\widehat{\mu}\pm\lambda\widehat{\sigma}$ for each component, where $\widehat{\mu}$ and $\widehat{\sigma}$ are estimated from the training set, and its novelty score is the number of components of $L(w)$ that exceed their confidence intervals.
\end{itemize}
Each of these methods produces a novelty score and, similarly to ours, requires a decision threshold computed on a small part of the dataset, as illustrated in Section \ref{subsec:open-set}. However, we evaluate the novelty-detection performance by the AUC, which does not depend on the threshold, nor, most importantly, on class proportions. SoftMax, PreSoftMax and SME do not require training other than that of the SSCN. In OpenMax, we follow the procedure presented in \cite{bendale2016towards} to compute the MAVs and fit the Weibull model from the entire dataset used to train the SSCN. For a fair comparison with our solution, we adopt in IFOR and CI the same procedure illustrated in Section \ref{subsec:open-set}, so we fit the Isolation Forest and compute the parameters $\widehat{\mu}$, $\widehat{\sigma}$ from a small portion of the dataset (5\%) that was not used to train the SSCN.

We remark that open-set recognition solutions based on the reconstruction error of autoencoders \cite{yoshihashi2019classification, oza2019c2ae, sun2020conditional} cannot be directly used on WDMs, hence we have not considered them. Instead, we have focused on open-set recognition methods built on top of a pre-trained SSCN, which can accurately classify WDMs from known classes. This enables a fair comparison between the novelty detection power of methods that achieve the same classification performance over known classes, which is of paramount importance in our industrial scenario.

\noindent\textbf{Results and discussion.} Table \ref{tab:ad_auc} reports the AUC achieved by the considered methods in novelty detection. As suggested in \cite{demvsar2006statistical}, we also report the average rank of the considered methods over the twelve \emph{Novel} classes and the p-values of the one-sided Wilcoxon Signed-Rank test~\cite{wilcoxon1945individual} assessing whether the difference between the AUC of our solution, which is the best in terms of ranking, and those of the other methods is significant. Moreover, we apply the nonparametric Mann-Whitney test on the novelty scores of the best and second-best performing methods on each \emph{Novel} class to assess whether the difference is statistically significant. This test is suggested in \cite{delong1988comparing} to compare the AUC of two methods on a binary-classification problem. 
\begin{table}[t!]
    \caption{Area Under the ROC Curve (AUC) of the considered methods in detecting the considered \emph{Novel} classes. Bold indicates the best values, and the underlined best AUC values are significantly higher than the second-best on the same novel class (p-value $\leq0.05$) according to the Mann-Whitney test \cite{delong1988comparing}. To compare the overall performance of the considered methods, we report the average rank over the novel classes and the p-values obtained by the one-sided Wilcoxon Signed-Rank test assessing the significance of the performance differences between our solution and each alternative method singularly, as recommended in \cite{demvsar2006statistical}.}
	\label{tab:ad_auc}
	\centering
	\small
	\vspace{0.1cm}
	\resizebox{\textwidth}{!}{
	\begin{tabular}{l|ccccccc}
        Novel class & SoftMax \cite{hendrycks2016baseline} & PreSoftMax & OpenMax \cite{bendale2016towards} & SME \cite{geng2020guided} & IFOR \cite{zhu2018multi} & CI \cite{zhang2020large} & GMM (ours)\\
        \hline
        BasketBall & 0.3697 & 0.3234 & 0.3455 & 0.3191 & 0.9812 & 0.4009 & $\textbf{\underline{0.9894}}$\\
        ClusterBig & 0.7022 & 0.8272 & 0.7216 & 0.6776 & 0.9541 & 0.6195 & $\textbf{0.9543}$\\
        ClusterSmall & 0.8672 & $\textbf{\underline{0.9003}}$ & 0.8775 & 0.8813 & 0.7694 & 0.8169 & 0.8227\\
        Donut & 0.4692 & 0.8508 & 0.6614 & 0.4270 & $\textbf{\underline{0.9596}}$ & 0.4916 & 0.9515\\
        Fingerprints & 0.8418 & 0.9222 & 0.8556 & 0.8234 & 0.8710 & 0.8214 & $\textbf{0.9249}$\\
        GeoScratch & 0.6377 & 0.7282 & 0.6699 & 0.6177 & 0.7997 & 0.8557 & $\textbf{0.8562}$\\
        Grid & 0.6716 & 0.5297 & 0.5454 & 0.6438 & $\textbf{0.8933}$ & 0.4989 & 0.8907\\
        HalfMoon & 0.7465 & 0.8566 & 0.7918 & 0.7375 & 0.8234 & 0.8641 & $\textbf{\underline{0.8873}}$\\
        Incomplete & 0.8162 & $\textbf{\underline{0.8438}}$ & 0.8319 & 0.8258 & 0.5752 & 0.7672 & 0.7244\\
        Ring & 0.4898 & 0.4928 & 0.5211 & 0.4626 & 0.8590 & 0.7376 & $\textbf{\underline{0.9196}}$\\
        Slice & 0.8313 & 0.6944 & 0.7772 & 0.7723 & 0.8887 & 0.7969 & $\textbf{\underline{0.9050}}$\\
        ZigZag & 0.6448 & 0.7251 & 0.6647 & 0.6269 & 0.8802 & 0.8865 & $\textbf{\underline{0.9149}}$\\
        \hline
        Avg. rank & 4.8333 & 3.7500 & 4.2500 & 5.7500 & 3.0833 & 4.4167 & $\textbf{1.9167}$\\
        Wilcoxon p & 0.0024 & 0.0061 & 0.0024 & 0.0012 & 0.0024 & 0.0024 & \textbf{--}\\
	\end{tabular}
	}
	\vspace{-0.3cm}
\end{table}

Table \ref{tab:ad_auc} indicates that the proposed GMM is the best method in detecting eight out of the twelve novel classes, five of which with statistical significance (p-value $\leq0.05$) according to the Mann-Whitney test \cite{delong1988comparing}. As we note in our first experiment, some classes are very similar and might be easily confused by the classifier (e.g., \emph{ClusterSmall} and \emph{Incomplete}), therefore they are more difficult to detect as novelties than others, as is apparent from Table \ref{tab:ad_auc}. Most importantly, our solution achieves the best average rank among the considered methods, and the p-values of the Wilcoxon test show that there is enough statistical evidence to claim that GMM performs significantly better (p-value $\leq0.05$) than all the alternative methods.


The AUC values achieved by IFOR follow a trend similar to those of GMM, even though inferior in most cases. Since we expect that $L(w)$ follows a multimodal distribution due to the presence of different known classes, the explicitly multimodal GMM outperforms IFOR, which is completely nonparametric. In contrast, SME and CI achieve relatively good results only on few specific novel classes, such as \emph{ClusterSmall}. 
SoftMax performs worse than most of the other methods: as expected, novelties are often classified with high confidence due to the SoftMax operation. OpenMax performs better than SoftMax according to the average ranks but worse than GMM. Perhaps surprisingly, PreSoftMax performs very well, and represents the best method to detect \emph{ClusterSmall} and \emph{Incomplete}. This is probably due to the fact that these two classes are difficult to distinguish, hence the SSCN often classifies them with low confidence.

The fact that PreSoftMax outperforms OpenMax and SoftMax outperforms SME shows that the highest classification score (before and after applying SoftMax) is the most informative for novelty detection in WDMs, confirming the intuition that novelties are likely to be classified with low confidence. By also considering the other scores, OpenMax and SME search for outliers in a 12-dimensional space where most dimensions are not informative for novelty detection, while PreSoftMax and SoftMax operate on a 1-dimensional space since they only consider the highest score. For this reason, we speculate that OpenMax and SME might suffer from an effect similar to \emph{detectability loss} \cite{alippi2016change} i.e., the higher the dimensionality, the harder it is to detect a distribution change.

\section{Conclusions}\label{sec:conclusion}
The effective and automatic monitoring of large volumes of silicon wafers is a crucial challenge to improve the quality and efficiency of industries operating in this sector. This paper addresses the problem of open-set pattern recognition on Wafer Defect Maps. We show that a simple deep-learning model based on Submanifold Sparse Convolutions taking full-resolution WDMs as input is substantially more robust to class imbalance compared to fine-tuning pre-trained CNNs, which can only process relatively small images. This suggests that binning WDMs to reduce their size leads to a relevant information loss, and thus Submanifold Sparse Convolutional Networks are the perfect instrument to process WDMs at their original resolution. Moreover, we are the first to address the open-set recognition problem on WDMs, which is of paramount importance for semiconductor industries because novel patterns might occur due to production issues that have never been observed before. To this purpose, we combine our Submanifold Sparse Convolutional network with a novelty detector based on a Gaussian Mixture Model fitted on the latent representation of the network. We show that our solution outperforms state-of-the-art open-set recognition methods in novelty detection. Moreover, we present a class-specific data augmentation procedure for WDMs that is fundamental to prevent the class imbalance from influencing the classification performance. We also show that our data augmentation can be safely used at test time because the transformations preserve the known class labels and the \emph{Novel} class label.


Besides the deployment of our solution in several STMicroelectronics production sites, we are studying other industrial settings that operate with point clouds, such as those described in \cite{bergamasco2020cylinders}. Moreover, we are investigating strategies for training at the same time the classifier and the GMM using the technique proposed in \cite{zong2018deep} to encourage the latent representation of an autoencoder to follow a Gaussian Mixture distribution.

\section*{Acknowledgements}
\vspace{-0.1cm}

We gratefully acknowledge the support of NVIDIA Corporation with the donation of the Titan Xp GPU that researchers from Politecnico di Milano have used in this work.

\bibliography{Bea}
\vspace{0.5cm}
\noindent\textbf{Luca Frittoli} graduated in Mathematics at  Universit\`a degli Studi di Milano in 2018, and is currently working towards the Ph.D. in Information Technology at Politecnico di Milano. His research interests regard change detection in datastreams and anomaly detection in non-matrix data.

\noindent\textbf{Diego Carrera} graduated in Mathematics at Universit\`a degli Studi di Milano in 2013 and received the Ph.D. in Information Technology at Politecnico di Milano in 2018. He joined STMicroelectronics in 2019, and he is currently developing automatic quality inspection systems to monitor the wafer production.

\noindent\textbf{Beatrice Rossi} graduated in Mathematics and Applications at Universit\`a degli Studi di Milano Bicocca in 2008. Since then she has been working at STMicroelectronics, System Research and Applications. Her research interests include Machine Learning techniques for Anomaly Detection and Image Classification, and Distributed Ledger Technology for the IoT.

\noindent\textbf{Pasqualina Fragneto} graduated in Mathematics at Universit\`a degli Studi di Napoli in 1998. Currently, she is leading the Applied Math Team at STMicroelectronics, System Research and Applications. Her activities concern the introduction of intelligent systems in production and the development of advanced and modular manufacturing technologies.

\noindent\textbf{Giacomo Boracchi} is an Associate Professor at Politecnico di Milano, where he also received the Ph.D. in Information Technology (2008). His research interests concern image processing and machine learning, and in particular algorithms for image restoration and analysis, change/anomaly detection, domain adaptation. 

\end{document}